# Samba: Semantic Segmentation of Remotely Sensed Images with State Space Model

Qinfeng Zhu, *Graduate Student Member, IEEE,* Yuanzhi Cai, *Member, IEEE,* Yuan Fang, *Graduate Student Member,* Yihan Yang, Cheng Chen, Lei Fan, *Member, IEEE,* Anh Nguyen

*Abstract*—High-resolution remotely sensed images pose a challenge for commonly used semantic segmentation methods such as Convolutional Neural Network (CNN) and Vision Transformer (ViT). CNN-based methods struggle with handling such high-resolution images due to their limited receptive field, while ViT faces challenges in handling long sequences. Inspired by Mamba, which adopts a State Space Model (SSM) to efficiently capture global semantic information, we propose a semantic segmentation framework for high-resolution remotely sensed images, named Samba. Samba utilizes an encoder-decoder architecture, with Samba blocks serving as the encoder for efficient multi-level semantic information extraction, and UperNet functioning as the decoder. We evaluate Samba on the LoveDA, ISPRS Vaihingen, and ISPRS Potsdam datasets, comparing its performance against top-performing CNN and ViT methods. The results reveal that Samba achieved unparalleled performance on commonly used remote sensing datasets for semantic segmentation. Our proposed Samba demonstrates the first time the effectiveness of SSM in semantic segmentation of remotely sensed images, setting a new benchmark in performance for Mamba-based techniques in this specific application. The source code and baseline implementations are available at https://github.com/zhuqinfeng1999/Samba.

Under Review. This work was supported in part by the Xi'an Jiaotong-Liverpool University Research Enhancement Fund under Grant REF-21-01-003, and in part by the Xi'an Jiaotong-Liverpool University Postgraduate Research Scholarship under Grant FOS2210JJ03. *(Corresponding author: Lei Fan.)*

Qinfeng Zhu is with the Department of Civil Engineering, Xi'an Jiaotong-Liverpool University, Suzhou, 215123, China, and also with the Department of Computer Science, University of Liverpool, Liverpool, L69 3BX, UK. (e-mail: Qinfeng.Zhu21@student.xjtlu.edu.cn)

Yuanzhi Cai is with the CSIRO Mineral Resources, Kesington, WA 6151, Australia. (e-mail: Yuanzhi.Cai@CSIRO.AU)

Yuan Fang, Cheng Chen, and Lei Fan are with the Department of Civil Engineering, Xi'an Jiaotong-Liverpool University, Suzhou, 215123, China. (e-mail: Yuan.Fang16@student.xjtlu.edu.cn; Cheng.Chen19@student.xjtlu.edu.cn; Lei.Fan@xjtlu.edu.cn)

Yihan Yang is with the Department of Electrical and Electronic Engineering, Xi'an Jiaotong-Liverpool University, Suzhou, 215123, China (e-mail: Yihan.Yang2102@student.xjtlu.edu.cn)

Anh Nguyen is with the Department of Computer Science, University of Liverpool, Liverpool, L69 3BX, UK (e-mail: Anh.Nguyen@liverpool.ac.uk)

Color versions of one or more of the figures in this article are available online at http://ieeexplore.ieee.org

## I. INTRODUCTION

SEMANTIC segmentation of remotely sensed images is a crucial task in many remote sensing applications, widely implemented using deep learning methods [1]. Among these, a commonly used deep learning technique is Convolutional Neural Network (CNN) [2-5]. By performing convolution operations that slide over image data, CNN effectively extracts semantic features from shallow to deep layers in images, serving as a cornerstone in numerous image processing tasks [6-8]. However, the limited receptive field in CNN presents a challenge, particularly in handling high-resolution images [9], as shown in Fig. 1.(a). Although solutions exist to mitigate this issue, they come with their own flaws. For example, while scaling images can adapt to the receptive field, it often results in resolution loss, which affects model performance. Dilated convolution can expand the receptive field, but it may lead to information loss [10, 11] because of coarse feature sub-sampling. Alternatively, connecting multiple CNNs through residual connections allows for integrating high-level semantics with low-level information, thereby enhancing the model's ability to recognize different scales [12, 13]. However, deep residual connections significantly increase network computation complexity.

Vision Transformer (ViT) [14] is another widely employed deep learning technique for semantic segmentation [15, 16]. With its global attention mechanism, ViT overcomes the limitation posed by the receptive field and is capable of flexibly adapting to inputs of varying resolutions [17], as shown in Fig. 1.(b). ViT has demonstrated remarkable performance in traditional image tasks [18, 19], such as classification in ImageNet dataset [6]. However, challenges remain when employing ViT semantic segmentation of remotely sensed images. The computational complexity increases exponentially with increasing resolution due to the need to compute the attention mechanism between each image patch. Furthermore, ViT requires a large amount of training data, which may be scarce in the context of remotely sensed imagery.

Recently, a novel approach called Mamba has been proposed, which utilizes a State Space Model (SSM) to capture global semantic information with low computational complexity [20]. Unlike Transformers, Mamba exhibits linear complexity, providing it with a distinct advantage in processing long sequences. It's interesting to explore the effect of replacing multi-head self-attention with Mamba in vision tasks.

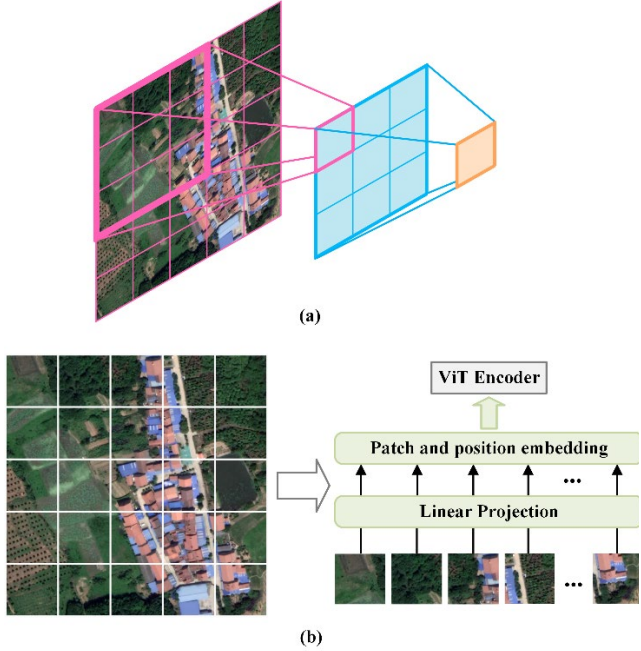

**Fig. 1.** (a) CNN's receptive field, which becomes 7×7 after two 3×3 convolutions, (b) ViT slices an image into several patches, and after linear projection and embedding, multi-head self-attention calculations are performed in the encoder to possess a global receptive field.

Inspired by Mamba, we propose Samba, a semantic segmentation framework tailored for remotely sensed images. The Samba block is specifically designed for efficient image feature extraction. In this framework, Mamba replaces multi-head self-attention in ViT to capture information from image data and is combined with multiple multi-layer perceptrons (MLP) to create a Samba block. The proposed semantic segmentation framework utilizes an encoder-decoder architecture, employing Samba blocks as the encoder and UperNet [21] as the decoder, to effectively extract multi-level semantic information.

The performance of our approach is evaluated using the LoveDA [22], ISPRS Vaihingen, and ISPRS Potsdam datasets. Comparing it against top-performing CNN and ViT methods, without loading pretrained parameters, Samba has achieved unparalleled performance on these datasets. This represents that Samba is an effective application of the State Space Model in semantic segmentation of remotely sensed images, establishing a benchmark in performance for Mamba-based techniques in this field.

The main contributions of this study can be summarized as follows:

1) We propose the Samba architecture, introducing the Mamba architecture into segmentation of remotely sensed images for the *first* time.
2) We conducted extensively comparative experiments against top-performing models, showcasing great potential of the Mamba architecture as a backbone for semantic segmentation of remotely sensed images.

3) We have established a new benchmark in performance for Mamba-based segmentation of remotely sensed images, and provided insights and potential directions for future work.

## II. METHODOLOGY

### A. Architecture Overview

Fig. 2. illustrates the encoder architecture of Samba, featuring Samba Blocks structured into four stages for progressive downsampling. Starting with an input image of dimensions $H \times W \times 3$, each Samba Block stage successively reduce its dimensions to $\frac{H}{4} \times \frac{W}{4} \times C$, $\frac{H}{8} \times \frac{W}{8} \times 2C$, $\frac{H}{16} \times \frac{W}{16} \times 4C$, and finally $\frac{H}{32} \times \frac{W}{32} \times 8C$. These progressively reduced features are subsequently processed by the UperNet decoder, which incrementally upsamples them to produce segmentation results.

### B. Samba Block

The ViT encoder uses multi-head self-attention to capture information within different representational subspaces, followed by residual connections and Layer Normalization (LN) to mitigate gradient vanishing. Subsequently, a Feed-Forward Network (FFN), consisting of an MLP and LN, is employed to introduce non-linear transformations and integrate complex information refined by multi-head self-attention.

Inspired by the robust architecture of the ViT encoder, the Samba Block adopts a similar architecture by replacing multi-head self-attention with a Mamba block. This Mamba block is utilized for feature extraction from high-resolution image sequences, avoiding quadratic complexity in computation. In our method, a combination of the Mamba Block and MLP is adopted to enhance the model's representational capacity and strengthen its learning ability for complex data.

### C. Mamba Block

Mamba enables SSM parameters to be functions of the input, facilitating the model to selectively propagate or discard information based on the current token. Therefore, it has attracted attention within the computer vision domain.

The core state space model of Mamba can be expressed by linear ordinary differential equations with evolution parameter $A$, and projection parameters $B$ and $C$:

$$h'(t) = Ah(t) + Bx(t) \tag{1}$$

$$y(t) = Ch(t) + Dx(t) \tag{2}$$

where $x(t)$ represents the input sequence, $h(t)$ represents the latent state, $h'(t)$ represents the update of the latent state, and $y(t)$ represents the predicted output sequence.

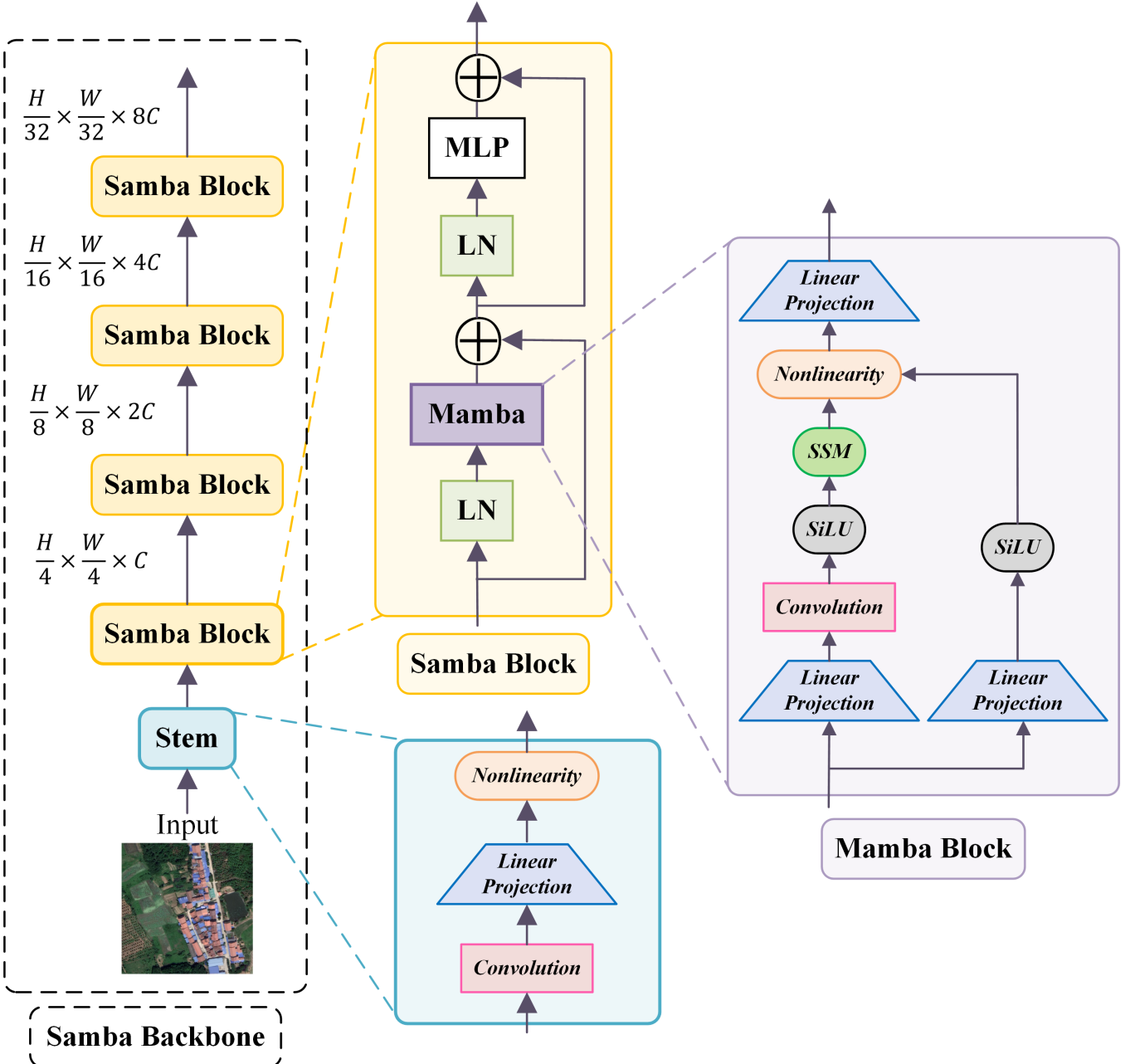

**Fig. 2.** Overall architecture of Samba.

The SSM maps $x(t)$ to the response $y(t)$ through the latent space of $h(t)$. In deep learning models, the required state transition is discrete rather than continuous. Therefore, the discretization of this state is important. The discrete outputs are obtained from the sampling values based on the time step of the input results. (1) and (2) can be discretized using (3) – (7):

$$h_k = \bar{A}h_{k-1} + \bar{B}x_k \tag{3}$$

$$y_k = \bar{C}h_k + \bar{D}x_k \tag{4}$$

$$\bar{A} = e^{\Delta A} \tag{5}$$

$$\bar{B} = (e^{\Delta A} - I)A^{-1}B \tag{6}$$

$$\bar{C} = C \tag{7}$$

where $\bar{A}$ and $\bar{B}$ represent the discretized matrix $A$ and $B$. $h_{k-1}$ represents the state of the previous timestep, and $h_k$ represents the state of the current timestep.

TABLE I
TRAINING SETTINGS FOR SEMANTIC SEGMENTATION NETWORKS.

| Decoder | Encoder | Total training iterations | Batch size | Optimizer | Initial learning rate | Warmup iterations | Learning rate schedule | Weight decay | Data augmentation |
|---|---|---|---|---|---|---|---|---|---|
| DeepLabV3+ | ResNet50 | 15k | 16 | SGD | 0.01 | NaN | PolyLR | 0.0005 | Random resize, Random Crop, Random Flip, Photo Metric Distortion |
| PSPNet | ResNet50 | | | | | | | | |
| UperNet | ResNet50 | | | | | | | | |
| UperNet | ConvNeXt | | | AdamW | 0.0001 | 1500 | | 0.05 | |
| UperNet | Swin-T | | | | 0.000006 | | | 0.1 | |
| Segformer | Mix ViT | 160k | | | 0.00006 | | | 0.01 | |
| UperNet | Samba | 15k | | | 0.0006 | | | 0.01 | |

Originally designed for one-dimensional sequences in natural language processing, Mamba encountered challenges when applied to the visual domain. Although some efforts [23, 24] attempted to introduce Mamba to visual tasks by converting images into unfolded two-dimensional patches and scanning them in multiple directions, tests revealed that its performance still lagged behind state-of-the-art ViT methods. In our proposed architecture, we do not scan the image patches in this way, and we linearly project the image's flattened patches in a manner similar to ViT. Specifically, we utilize a combination of convolutional layers, projection layers, and layer normalization for the input of the Mamba Block. This block is constructed based on the H3 architecture [25], a well-known SSM architecture, combined with a Gated MLP, as shown on the right side of Fig. 2. The sequence length of flattened patches derived from high-resolution remote sensing imagery is substantial. Mamba models this extensive sequence, endowing each patch with semantic information induced from the contextual sequence. Consequently, this approach facilitates the effective extraction of features from high-resolution remote sensing images.

## III. EXPERIMENTS

### A. Datasets and Metrics

Our proposed framework is validated using three commonly used benchmark datasets, including LoveDA, ISPRS Vaihingen, and Potsdam datasets, as detailed as follows.

LoveDA consists of remotely sensed RGB imagery with a spatial resolution of 0.3 meters, including 7 semantic classes. Its image resolution is $1024 \times 1024$. It consists of 2522 training images, 1669 validation images, and 1796 test images, with the validation set used for performance evaluation.

ISPRS Vaihingen consists of remotely sensed imagery with a spatial resolution of 9 centimetres, and includes 6 semantic classes. Its average image resolution is $2494 \times 2064$. Near infrared (IR), red, and green channels are used in our experiment. We use tiles 2, 4, 6, 8, 10, 12, 14, 16, 20, 22, 24, 27, 29, 31, 33, 35, 38 for testing (the same set of images for validation too) in this study, with the remaining images allocated for training. The clutter class is not calculated in evaluation.

ISPRS Potsdam consists of remotely sensed imagery with a spatial resolution of 5 centimetres, featuring 6 semantic classes. Each image has a resolution is $6000 \times 6000$. Red, green, and blue (RGB) channels are used in our experiments. For testing and validation, we use images with ID 2_13, 2_14, 3_13, 3_14, 4_13, 4_14, 4_15, 5_13, 5_14, 5_15, 6_13, 6_14, 6_15, 7_13 in this study. Training is implemented using the remaining images, following the same setting as SBSS [5]. The clutter class is not calculated in evaluation.

The Mean Intersection over Union (mIoU) is used to assess the accuracy of segmentation, and is calculated using Eq. (8):

$$\text{mIoU} = \frac{1}{C}\sum_{c=1}^{C}\frac{TP_c}{TP_c+FP_c+FN_c} \quad (8)$$

where $TP_c$, $FP_c$, and $FN_c$ respectively represent the true positives, false positives, and false negatives for class $c$, elaborated as follows:

True Positives ($TP_c$): These are the pixels or areas correctly identified as belonging to class $c$. It means that both the predicted label and the true label agree on class $c$.

False Positives ($FP_c$): These refer to the pixels or areas incorrectly labeled as class $c$ by the model, but in reality, they belong to a different class. This error type reflects overestimation of class $c$ presence.

False Negatives ($FN_c$): These are the pixels or areas that truly belong to class $c$ but were missed or incorrectly labeled as another class by the model. This represents an underestimation of class $c$ 's presence.

### B. Training Settings

In this study, we evaluate our method against several established approaches known for their effectiveness. These include CNN-based methods such as ConvNeXt [26], ResNet [27], Deeplab V3+ [12], and PSPNet [28], and ViT-based methods such as Swin-T [29], and Segformer [30]. To ensure fair comparisons, the tested methods are uniformly initialized without their pre-trained parameters, and input images are cropped into 512×512. The optimization and learning rate strategy settings for these methods adhere to widely adopted optimal configurations. Cross entropy is used as the loss function. Data augmentation can effectively enhance the

TABLE II

ACCURACY OF SEMANTIC SEGMENTATION ON THE LoveDA DATASET FROM SAMBA AND OTHER COMPARED METHODS. THE ACCURACY OF EACH CATEGORY IS PRESENTED BY THE IoU METRIC. THE HIGHEST SCORES ARE HIGHLIGHTED IN BOLD.

| Decoder | Encoder | mIoU | Background | Building | Road | Water | Barren | Forest | Agricultural |
|---------|---------|------|------------|----------|------|-------|--------|--------|--------------|
| UperNet | ConvNeXt | 36.81 | 45.51 | 30.48 | 43.53 | 49.17 | 17.09 | 35.19 | 36.70 |
| UperNet | ResNet50 | 32.86 | 32.23 | 48.15 | 38.97 | 37.66 | 14.17 | 19.26 | 39.58 |
| DeepLab V3+ | ResNet50 | 34.6 | 43.55 | 47.32 | 41.17 | 39.41 | 22.43 | 20.69 | 36.65 |
| PSPNet | ResNet50 | 33.73 | 36.3 | 48.81 | 35.03 | 36.34 | 21.10 | 23.74 | 34.82 |
| UperNet | Swin-T | 41.08 | 49.78 | 49.58 | 40.58 | 52.12 | 22.61 | 34.87 | 37.98 |
| Segformer | Mix ViT | 43.16 | 50.71 | 49.89 | 46.12 | 47.79 | **23.95** | **41.03** | 42.60 |
| UperNet | **Samba(Ours)** | **47.11(1st)** | **51.71** | **57.85** | **49.86** | **59.85** | 21.93 | 41.00 | **47.56** |

TABLE III

ACCURACY OF SEMANTIC SEGMENTATION ON THE ISPRS VAIHINGEN DATASET FROM SAMBA AND OTHER COMPARED METHODS. THE ACCURACY OF EACH CATEGORY IS PRESENTED BY THE IoU METRIC. THE HIGHEST SCORES ARE HIGHLIGHTED IN BOLD.

| Decoder | Encoder | mIoU | Impervious surface | Building | Low vegetation | Tree | Car |
|---------|---------|------|--------------------|----------|----------------|------|-----|
| UperNet | ConvNeXt | 67.42 | 76.89 | 81.71 | 60.48 | 73.11 | 44.01 |
| UperNet | ResNet50 | 70.25 | 79.18 | 83.35 | 65.76 | 77.48 | 45.52 |
| DeepLab V3+ | ResNet50 | 69.1 | 77.89 | 82.63 | 64.85 | 76.76 | 43.38 |
| PSPNet | ResNet50 | 68.72 | 78.80 | 82.84 | 65.54 | 77.11 | 39.30 |
| UperNet | Swin-T | 71.72 | 80.43 | 85.18 | **67.30** | 77.69 | 48.01 |
| Segformer | Mix ViT | 70.23 | 79.38 | 85.77 | 64.13 | 75.87 | 46.01 |
| UperNet | **Samba(Ours)** | **73.56(1st)** | **81.67** | **87.26** | 65.82 | **77.93** | **55.10** |

TABLE IV

ACCURACY OF SEMANTIC SEGMENTATION ON THE ISPRS POTSDAM DATASET FROM SAMBA AND OTHER COMPARED METHODS. THE ACCURACY OF EACH CATEGORY IS PRESENTED BY THE IoU METRIC. THE HIGHEST SCORES ARE HIGHLIGHTED IN BOLD.

| Decoder | Encoder | mIoU | Impervious surface | Building | Low vegetation | Tree | Car |
|---------|---------|------|--------------------|----------|----------------|------|-----|
| UperNet | ConvNeXt | 74.7 | 79.57 | 85.64 | 66.79 | 62.09 | 79.42 |
| UperNet | ResNet50 | 74.98 | 78.11 | 82.52 | 68.97 | 65.87 | 79.42 |
| DeepLab V3+ | ResNet50 | 75.23 | 78.37 | 82.76 | 68.32 | 66.54 | 80.18 |
| PSPNet | ResNet50 | 73.13 | 77.49 | 82.02 | 67.10 | 64.47 | 74.47 |
| UperNet | Swin-T | 76.46 | 80.97 | 86.45 | 70.20 | 66.03 | 78.67 |
| Segformer | Mix ViT | 81.13 | 83.40 | 89.63 | 72.74 | 73.38 | 86.51 |
| UperNet | **Samba(Ours)** | **82.29(1st)** | **84.45** | **90.06** | **74.37** | **74.98** | **87.61** |

generalization ability of deep learning models when training data is limited [31], so we use random resize, random crop, random flip, and photo metric distortion to augment training data. Specific training settings are summarized in Table I. All experiments are conducted using two NVIDIA RTX 3090 GPUs (24G) and two 4090D GPUs (24G).

## C. Results

The experimental results are summarized in Table II, III, and IV, highlighting the performance of Samba against the current top-performing CNN-based and ViT-based methods. Samba achieved the best performance on the LoveDA, ISPRS Vaihingen, and ISPRS Potsdam datasets, significantly surpassing CNN-based methods and slightly exceeding the ViT-based methods. Specifically for LoveDA, when using UperNet as the decoder, Samba outperformed the best-performing ViT-based model Segformer by 3.95% in the mIoU metric, and surpassed the best-performing CNN-based model ConvNeXt by 10.3% in terms of the mIoU metric. Specifically, Samba demonstrated the most significant improvements in the Building, Water, and Agricultural categories, with improvements of 7.96%, 7.73%, and 4.96%, respectively, compared to the most effective method among those compared.

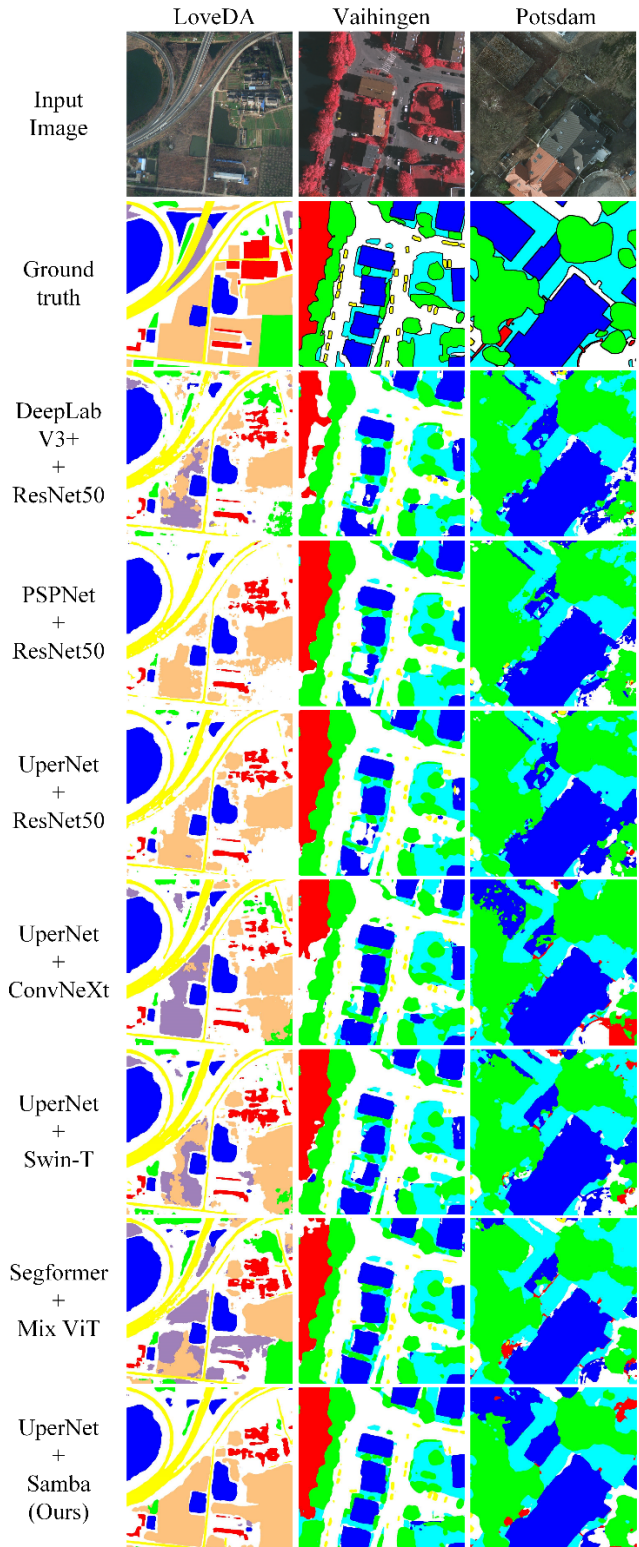

**Fig. 3.** Visual comparisons of segmentation outcomes between Samba and other compared methods.

TABLE V
COMPARISON OF COMPUTATIONAL COMPLEXITY AND PARAMETERS BETWEEN SAMBA AND OTHER COMPARED METHODS.

| Decoder | Encoder | Flops per patch (G) | Parameters (M) |
|---------|---------|---------------------|----------------|
| UperNet | ConvNeXt | 234 | 59.2 |
| UperNet | ResNet50 | 237 | 64.0 |
| DeepLab V3+ | ResNet50 | 177 | 41.2 |
| PSPNet | ResNet50 | 179 | 46.6 |
| UperNet | Swin-T | 236 | 58.9 |
| Segformer | Mix ViT | - | - |
| UperNet | **Samba (Ours)** | 232 | 51.9 |

Despite significant differences in training data quantities, spatial resolutions, and segmentation categories among the considered datasets, similar results were also achieved on the ISPRS Vaihingen and ISPRS Potsdam datasets. Samba surpassed other methods in terms of the mIoU metric as well as the IoU for most of the segmentation categories. When combined with the UperNet decoder, Samba utilized lower flops per patch and parameters than Swin-T, ResNet50, and ConvNeXt, as shown in Table V.

Moreover, as shown in Fig. 3., we visualized the semantic segmentation results achieved by Samba along with those compared models on the LoveDA, ISPRS Vaihingen, and ISPRS Potsdam dataset. These visualizations further demonstrate Samba's consistently enhanced performance in semantically segmenting high-resolution remotely sensed images.

## IV. DISCUSSION

The findings of this study highlight the effectiveness of Mamba in semantically segmenting high-resolution remotely sensed images, pushing forward state-of-the-art accuracy while demanding lower computational resources compared to other methods considered. Although the performance of Mamba approaches that of ViT in visual tasks, the newly proposed Samba outperforms ViT in semantic segmentation. This superior performance is attributed to the architecture of Mamba, which is exceptionally suited for this type of tasks, demonstrating its capability in efficiently capturing global semantic information.

Observations in Fig. 3. reveal that semantic segmentation results of CNN-based methods exhibit a notable dispersion across local regions, manifesting in a tendency towards fragmented segmentation. Additionally, segmentation results exhibit blurred segmentation boundaries, and incomplete segmentation of large-scale objects. This is due to the inherent limitation of CNNs with fixed receptive fields. Although adept at extracting local features, they struggle to effectively capture large-scale contextual information when dealing with high-resolution remotely sensed images, leading to a piecemeal recognition of features within the scene. Simultaneously, CNN-based methods frequently encounter false positive errors in large-scale objects due to their deficiency in understanding context. It underscores the challenge faced by CNN-based methods in achieving holistic and contiguous segmentation outcomes, particularly in complex remote sensing imagery where contextual coherence and spatial continuity are crucial for accurate land cover and land use classification. Compared to CNN-based methods, ViT-based methods exhibit smoother segmentation results but also suffer from the issue of blurred boundaries. Despite their advantage in understanding the global context, ViT sacrifices the fidelity and precision of capturing local features. The manifestation of false positives in localized areas underscores a critical challenge for ViT models in balancing the extraction and integration of global and local cues to achieve accurate and reliable segmentation when faced with high-resolution remotely sensed images.

Compared to the LoveDA dataset, which comprises 2522 training images, the training data available for the ISPRS Vaihingen and ISPRS Potsdam datasets are considerably limited. Both CNN-based and ViT-based methods exhibit a pronounced issue of FN errors in these datasets, especially in larger areas, such as the building class (indicated by blue in Fig. 3. for Vaihingen and Potsdam). Conversely, Samba manages to achieve relatively complete segmentation results even with limited training data. While CNN-based and ViT-based methods can produce relatively complete segmentation outcomes on the LoveDA dataset, they manifest significant FP errors in larger regions, for instance, the agricultural areas (represented by light apricot color in Fig. 3. of LoveDA) are categorized into barren areas (represented by the light purple color), whereas Samba seldom encounters such errors in the segmentation of large areas. The Agricultural category was both crucial and challenging to identify within the LoveDA dataset. Samba's exemplary performance in this category underscores the effectiveness of its architecture in extracting information through context induction in the semantic segmentation of high-resolution remote sensing imagery.

Thanks to the SSM's powerful inductive capability in long sequences, Samba exhibited outstanding segmentation results in the dataset considered. Samba delivered more complete and accurate segmentation of large terrain areas, compared to other methods, thus performing well in minimizing false positives. However, like ViT, it suffered from a lower focus on local details, leading to the omission of certain small objects and resulting in false negative errors. This observation points to the need for additional strategies to mitigate the impact of overgeneralized feature attribution on segmentation performance.

Based on our research, we suggest the following directions for future investigations.

1) Despite its advantage in long sequence induction capability and low computational complexity, the ability of Mamba-based methods to extract local information is limited. Future work could explore combining Mamba with CNNs to enhance the capability of capturing details.

2) Given the limited access to annotated remote sensing image data, transfer learning is considered an important technique for segmentation tasks. The pre-trained models can be obtained after training on large-scale datasets, such as ImageNet [6]. Exploring efficient and effective transfer learning methods tailored to the Mamba architecture is also potential research direction.

3) Since Mamba-based methods perform well in dealing with long sequences, it is valuable to explore their application in semantic segmentation of multi-channel data [32, 33], such as hyperspectral data.

4) In our experiments, we observed that the Mamba architecture occasionally exhibited instability when training from scratch, a common issue identified in other studies [34]. While this issue may be partially mitigated by pre-training, a holistic investigation into it is valuable.

## V. CONCLUSION

This article introduces Samba, a novel semantic segmentation framework built on Mamba, specifically designed for segmentation of high-resolution remotely sensed images, marking the first integration of Mamba within the domain.

By evaluating its performance on the LoveDA, ISPRS Vaihingen, and ISPRS Potsdam datasets, Samba surpassed stat-of-the-art CNN-based and ViT-based methods on all those three datasets. Particularly for LoveDA, with UperNet serving as the decoder, Samba exceeded the performance of Segformer (a leading ViT-based model) and ConvNeXt (a top-performing CNN-based model) by 3.95% and 10.3%, respectively, in terms of the mIoU metric. These set new benchmarks in performance and demonstrating the effectiveness and potential of the Mamba architecture in semantic segmentation of high-resolution remote sensing imagery.

## REFERENCES

[1] Y. Mo, Y. Wu, X. Yang, F. Liu, and Y. Liao, "Review the state-of-the-art technologies of semantic segmentation based on deep learning," *Neurocomputing*, vol. 493, pp. 626-646, 2022.

[2] X. Yuan, J. Shi, and L. Gu, "A review of deep learning methods for semantic segmentation of remote sensing imagery," *Expert Systems with Applications*, vol. 169, pp. 114417, 2021.

[3] O. Ronneberger, P. Fischer, and T. Brox, "U-net: Convolutional networks for biomedical image segmentation." pp. 234-241.

[4] J. Long, E. Shelhamer, and T. Darrell, "Fully convolutional networks for semantic segmentation." pp. 3431-3440.

[5] Y. Cai, L. Fan, and Y. Fang, "SBSS: Stacking-based semantic segmentation framework for very high-resolution remote sensing image," *IEEE Transactions on Geoscience and Remote Sensing*, vol. 61, pp. 1-14, 2023.

[6] A. Krizhevsky, I. Sutskever, and G. E. Hinton, "Imagenet classification with deep convolutional neural networks," *Advances in neural information processing systems*, vol. 25, 2012.

[7] Y. Cai, L. Fan, P. M. Atkinson, and C. Zhang, "Semantic segmentation of terrestrial laser scanning

point clouds using locally enhanced image-based geometric representations," *IEEE Transactions on Geoscience and Remote Sensing*, vol. 60, pp. 1-15, 2022.

[8] C. Chen, and L. Fan, "Scene segmentation of remotely sensed images with data augmentation using U-net++." pp. 201-205.

[9] G. Lin, A. Milan, C. Shen, and I. Reid, "Refinenet: Multi-path refinement networks for high-resolution semantic segmentation." pp. 1925-1934.

[10] F. Yu, and V. Koltun, "Multi-scale context aggregation by dilated convolutions," *arXiv preprint arXiv:1511.07122*, 2015.

[11] R. Hamaguchi, A. Fujita, K. Nemoto, T. Imaizumi, and S. Hikosaka, "Effective use of dilated convolutions for segmenting small object instances in remote sensing imagery." pp. 1442-1450.

[12] L.-C. Chen, G. Papandreou, I. Kokkinos, K. Murphy, and A. L. Yuille, "Deeplab: Semantic image segmentation with deep convolutional nets, atrous convolution, and fully connected crfs," *IEEE transactions on pattern analysis and machine intelligence*, vol. 40, no. 4, pp. 834-848, 2017.

[13] Q. Zeng, J. Zhou, and X. Niu, "Cross-scale feature propagation network for semantic segmentation of high-resolution remote sensing images," *IEEE Geoscience and Remote Sensing Letters*, 2023.

[14] A. Dosovitskiy, L. Beyer, A. Kolesnikov, D. Weissenborn, X. Zhai, T. Unterthiner, M. Dehghani, M. Minderer, G. Heigold, and S. Gelly, "An image is worth 16x16 words: Transformers for image recognition at scale," *arXiv preprint arXiv:2010.11929*, 2020.

[15] X. Zhou, L. Zhou, S. Gong, S. Zhong, W. Yan, and Y. Huang, "Swin Transformer Embedding Dual-Stream for Semantic Segmentation of Remote Sensing Imagery," *IEEE Journal of Selected Topics in Applied Earth Observations and Remote Sensing*, 2023.

[16] M. Yao, Y. Zhang, G. Liu, and D. Pang, "SSNet: A Novel Transformer and CNN Hybrid Network for Remote Sensing Semantic Segmentation," *IEEE Journal of Selected Topics in Applied Earth Observations and Remote Sensing*, 2024.

[17] K. Han, Y. Wang, H. Chen, X. Chen, J. Guo, Z. Liu, Y. Tang, A. Xiao, C. Xu, and Y. Xu, "A survey on vision transformer," *IEEE transactions on pattern analysis and machine intelligence*, vol. 45, no. 1, pp. 87-110, 2022.

[18] M. Wortsman, G. Ilharco, S. Y. Gadre, R. Roelofs, R. Gontijo-Lopes, A. S. Morcos, H. Namkoong, A. Farhadi, Y. Carmon, and S. Kornblith, "Model soups: averaging weights of multiple fine-tuned models improves accuracy without increasing inference time." pp. 23965-23998.

[19] X. Zhai, A. Kolesnikov, N. Houlsby, and L. Beyer, "Scaling vision transformers." pp. 12104-12113.

[20] A. Gu, and T. Dao, "Mamba: Linear-time sequence modeling with selective state spaces," *arXiv preprint arXiv:2312.00752*, 2023.

[21] T. Xiao, Y. Liu, B. Zhou, Y. Jiang, and J. Sun, "Unified perceptual parsing for scene understanding." pp. 418-434.

[22] J. Wang, Z. Zheng, A. Ma, X. Lu, and Y. Zhong, "LoveDA: A remote sensing land-cover dataset for domain adaptive semantic segmentation," *arXiv preprint arXiv:2110.08733*, 2021.

[23] L. Zhu, B. Liao, Q. Zhang, X. Wang, W. Liu, and X. Wang, "Vision mamba: Efficient visual representation learning with bidirectional state space model," *arXiv preprint arXiv:2401.09417*, 2024.

[24] Y. Liu, Y. Tian, Y. Zhao, H. Yu, L. Xie, Y. Wang, Q. Ye, and Y. Liu, "Vmamba: Visual state space model," *arXiv preprint arXiv:2401.10166*, 2024.

[25] D. Y. Fu, T. Dao, K. K. Saab, A. W. Thomas, A. Rudra, and C. Ré, "Hungry hungry hippos: Towards language modeling with state space models," *arXiv preprint arXiv:2212.14052*, 2022.

[26] Z. Liu, H. Mao, C.-Y. Wu, C. Feichtenhofer, T. Darrell, and S. Xie, "A convnet for the 2020s." pp. 11976-11986.

[27] K. He, X. Zhang, S. Ren, and J. Sun, "Deep residual learning for image recognition." pp. 770-778.

[28] H. Zhao, J. Shi, X. Qi, X. Wang, and J. Jia, "Pyramid scene parsing network." pp. 2881-2890.

[29] Z. Liu, Y. Lin, Y. Cao, H. Hu, Y. Wei, Z. Zhang, S. Lin, and B. Guo, "Swin transformer: Hierarchical vision transformer using shifted windows." pp. 10012-10022.

[30] E. Xie, W. Wang, Z. Yu, A. Anandkumar, J. M. Alvarez, and P. Luo, "SegFormer: Simple and efficient design for semantic segmentation with transformers," *Advances in neural information processing systems*, vol. 34, pp. 12077-12090, 2021.

[31] Q. Zhu, L. Fan, and N. Weng, "Advancements in point cloud data augmentation for deep learning: A survey," *arXiv preprint arXiv:2308.12113*, 2023.

[32] Y. Cai, H. Huang, K. Wang, C. Zhang, L. Fan, and F. Guo, "Selecting optimal combination of data channels for semantic segmentation in city information modelling (CIM)," *Remote Sensing*, vol. 13, no. 7, pp. 1367, 2021.

[33] Y. Cai, L. Fan, and C. Zhang, "Semantic segmentation of multispectral images via linear compression of bands: An experiment using RIT-18," *Remote Sensing*, vol. 14, no. 11, pp. 2673, 2022.

[34] B. N. Patro, and V. S. Agneeswaran, "SiMBA: Simplified Mamba-Based Architecture for Vision and Multivariate Time series," *arXiv preprint arXiv:2403.15360*, 2024.

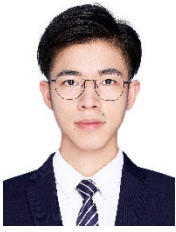

**Qinfeng Zhu** (Graduate Student Member, IEEE) received the degree of Master of Research in Computer Science from the University of Liverpool, Liverpool, U.K., in 2023, where he is currently working towards the Ph.D. degree in Computer Science.

His research interests mainly lie on deep learning, especially in multi-modal information fusion, 3D computer vision, semantic segmentation, and data augmentation.

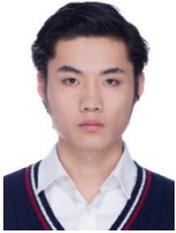

**Yuanzhi Cai** (Member, IEEE) received the Ph.D degree from the University of Liverpool, UK, in 2023.

He is currently a CERC Fellow within CSIRO Mineral Resources, Kesington, Australia. His research interests include the classification and segmentation of remote sensing data.

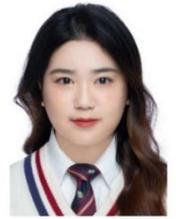

**Yuan Fang** (Graduate Student Member, IEEE) received the B.E. degree in civil engineering from the University of Liverpool, UK, in 2020. She is now working toward her PhD degree at the same University.

Her research interests include deep learning and data fusion of satellite images.

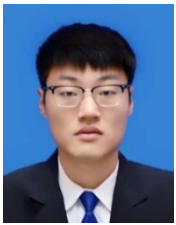

**Yihan Yang** received the master's degree in electronic communications and computer engineering from the University of Nottingham, UK, in 2021. He is currently working toward the Ph.D. degree in electrical and electronic engineering with the University of Liverpool, Liverpool, U.K.

His research area includes multi-view object tracking and detection, synthetic datasets and remote sensing.

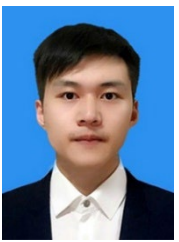

**Cheng Chen** received his B.E. degree in civil engineering from the Hefei University of Technology in 2016.

He is currently working on his Ph.D. program at the University of Liverpool, UK. His research areas are mainly focused on monitoring of geohazards, Interferometric Synthetic Aperture Radar and Geographic Information System.

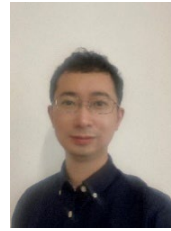

**Lei Fan** (Member, IEEE) received the Ph.D. degree from the University of Southampton, Southampton, UK, in 2018.

He is currently an Associate Professor within Department of Civil Engineering at Xi'an Jiaotong Liverpool University, Suzhou, China. His main research interests include lidar and photogrammetry techniques, point cloud, machine learning, deformation monitoring, semantic segmentations, monitoring of civil engineering structures and geohazards.

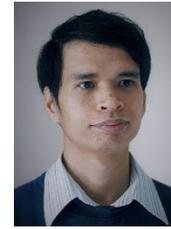

**Anh Nguyen** received the Ph.D. degree from the Italian Institute of Technology (IIT), Genoa, Italy, in 2019.

He worked at the Hamlyn Centre for Robotic Surgery, Imperial College London, London, U.K.; Australian National University, Canberra, ACT, Australia; National Institute for Research in Digital Science and Technology (Inria), France, and The University of Adelaide, Adelaide, SA, Australia. He is currently an Assistant Professor at the Department of Computer Science, University of Liverpool, Liverpool, U.K. His research interests include computer vision (image segmentation, detection, and medical imaging), machine learning (deep learning, deep reinforcement learning, and federated learning), and robotics (autonomous navigation/manipulation and medical robots).